\documentclass[letterpaper, 10 pt, conference]{ieeeconf}  

\IEEEoverridecommandlockouts                              

\usepackage{graphics} 
\usepackage{epsfig} 
\usepackage{mathptmx} 
\usepackage{times} 
\usepackage{amsmath} 
\usepackage{amssymb}  
\usepackage{url} 
\usepackage[hidelinks]{hyperref}
\usepackage{multirow} 
\usepackage{booktabs} 
\usepackage{caption}
\usepackage{graphicx}
\usepackage{subcaption}
\usepackage{calrsfs}
\usepackage{array} 
\DeclareMathAlphabet{\pazocal}{OMS}{zplm}{m}{n}
\SetMathAlphabet\pazocal{bold}{OMS}{zplm}{bx}{n}

\newcommand{\algoname}{STEPP}

\title{\LARGE \bf
Watch Your STEPP:\\Semantic Traversability Estimation using Pose Projected Features 
}

\author{Sebastian Ægidius, Dennis Hadjivelichkov, Jianhao Jiao, Jonathan Embley-Riches, and Dimitrios Kanoulas
\thanks{The authors are with the Robot Perception and Learning Lab, Department of Computer Science, University College London, Gower Street, WC1E 6BT, London, UK. {\tt\small \{s.aegidius, d.kanoulas\}@ucl.ac.uk}}
\thanks{Dimitrios Kanoulas is also with Archimedes/Athena RC, Greece.}
\thanks{This work was supported by the UKRI FLF [MR/V025333/1] (RoboHike).  For the purpose of Open Access, the author has applied a CC BY public copyright license to any Author Accepted Manuscript version arising from this submission.}}

\begin{document}

\maketitle
\thispagestyle{empty}
\pagestyle{empty}

\begin{abstract}
Understanding the traversability of terrain is essential for autonomous robot navigation, particularly in unstructured environments such as natural landscapes. Although traditional methods, such as occupancy mapping, provide a basic framework, they often fail to account for the complex mobility capabilities of some platforms such as legged robots. In this work, we propose a method for estimating terrain traversability by learning from demonstrations of human walking. Our approach leverages dense, pixel-wise feature embeddings generated using the DINOv2 vision Transformer model, which are processed through an encoder-decoder MLP architecture to analyze terrain segments. The averaged feature vectors, extracted from the masked regions of interest, are used to train the model in a reconstruction-based framework. By minimizing reconstruction loss, the network distinguishes between familiar terrain with a low reconstruction error and unfamiliar or hazardous terrain with a higher reconstruction error. This approach facilitates the detection of anomalies, allowing a legged robot to navigate more effectively through challenging terrain. We run real-world experiments on the ANYmal legged robot both indoor and outdoor to prove our proposed method. The code is open-source, while video demonstrations can be found on our website: \url{https://rpl-cs-ucl.github.io/STEPP/}
\end{abstract}

\section{INTRODUCTION}\label{Sec:intro}
Understanding the traversability of the surrounding terrain is essential for efficient navigation in robotics. Autonomous robots are required to choose safe and efficient routes when navigating challenging terrain and therefore must be able to judge the cost of traversability on surrounding terrain in some way. Navigation in unstructured environments in nature is challenging and traversability estimation can play a pivotal role.

To define a traversability cost to any observed terrain, an accurate understanding of what the robotic system is physically capable of is essential. What wheeled and flying robots can traverse over and through is a domain widely studied, but some platforms, such as legged robots with more advanced mobility skills, can enable navigation in more complex terrains. Classical approaches such as occupancy mapping~\cite{occupancymapping}, have long been the foundational idea of previous navigation systems and are often coupled with visual~\cite{jiao2025LiteVLoc} or LiDAR-enabled SLAM solutions to quickly understand the traversability of the surrounding terrain. Most often, the surrounding terrain is estimated on the measured height and distance from the open space. Such solutions are generally what are used for the standard local planner~\cite{DWA, TEB, Falco, liu2024dipper, yao2024local, ellis2022navigation} and often do not take into account the traversability abilities of a robot in unstructured terrain. 

\begin{figure}
    \centering
    \includegraphics[width=1.0\linewidth]{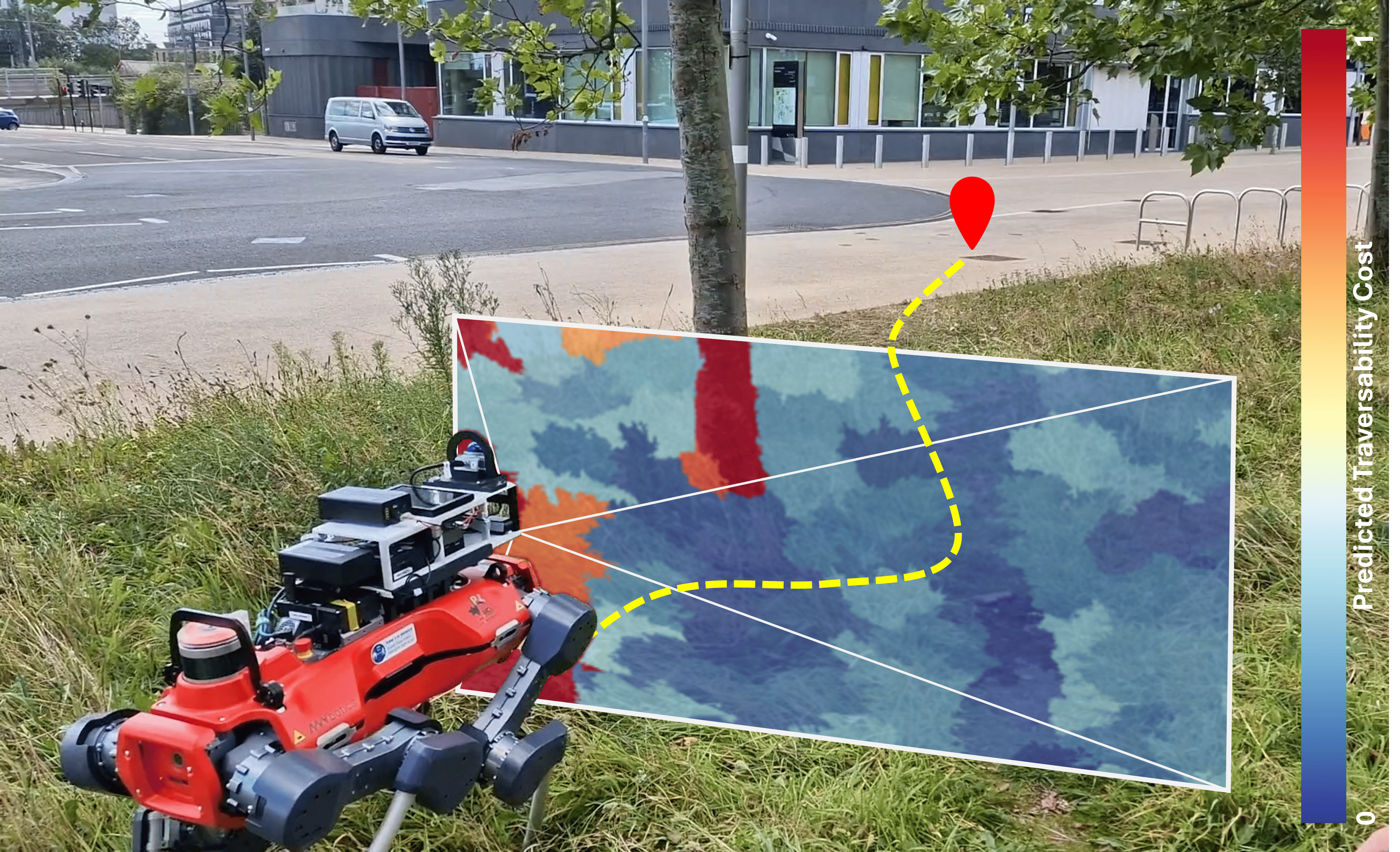}
    \caption{An illustration of \algoname\ with robot and traversability costs overlaid. Blue -- traversable, Red -- Untraversable.}
    \label{fig:1}
\end{figure}

Estimating terrain for navigation requires understanding both what is feasible and optimal. When humans or animals walk through a forest or over vegetation-covered terrain, they naturally prefer to walk on flatter ground or existing trails. Although walking on grass or plants is possible, choosing paths with shorter vegetation, trails, or paved roads reduces the risk of encountering obstacles such as stones, water, or mud hidden beneath plants. A more comprehensive understanding of the terrain and what is navigable for legged robots is essential. Recent works use supervised semantic segmentation learning to estimate traversable terrain~\cite{pmlr-v164-shaban22a, vstrong, vivekanandan2019terrain, walas2016terrain}. Others have proposed reinforcement learning methods that aim to model the environment and learn how to navigate it~\cite{BADGR, vint}. Although some approaches claim to be robot-agnostic, most remain specific to the robot used during training, limiting their adaptability during implementation and deployment. Working with data of humans directly navigating the world is a way to gain understanding of what is possible for legged robots, as well as how to maneuver through unstructured terrain.

In this work, we present \algoname, a robot-agnostic terrain estimation model capable of accurately and efficiently inferring the cost of navigating the surrounding terrain using only RGB input, without bias towards structured or unstructured terrain. \algoname\ generates a traversability costmap that can be utilized by a local planner to safely navigate the environment to a predetermined target point. The model has learned from data of humans walking through a variety of structured and unstructured terrain. Our research into traversability estimation is particularly inspired by the work of Frey et al.~\cite{wvn} and their recent breakthroughs in the use of high-dimensional features from pre-trained self-supervised models, integrated with real-time environmental feedback mechanisms. In this work, we aim to expand on the possibilities of further utilizing high-dimensional features from pre-trained foundation models. We allow navigation by extracting learnable features from offline data that capture humans walking across structured and unstructured terrains combined with synthetic generated data simulating walking across more varied terrain. We use an offline positive unlabeled data training approach to mitigate the difficulties of describing bad terrain to navigate over. The main contributions of this work are as follows. We introduce:
\begin{itemize}
    \item Feature interpretation for traversability estimation and a pre-trained model for indoor and outdoor navigation.
    \item A pipeline to integrate terrain estimation into a local path planner.
    \item Real-world experiments demonstrating live terrain estimation.
    \item Open-source code base for training and deployment and the Unreal Engine plugin for generating synthetic future path projection data in simulation.
\end{itemize}

\section{RELATED WORK}\label{Sec:background}
Classical approaches to estimating navigation costs are based primarily on geometric analysis of the surroundings using 3D information.They often use metric such as step difficulty and risk to determine traversability~\cite{step, tare}. 
Geometric approaches are limited because the geometric data do not indicate what the data represent, which is especially problematic in unstructured environments, where height maps and plane curvatures become impractical. Lorenz et al.~\cite{rough} allowed a quadruped to traverse rough terrain using a geometric approach to foothold planning. Similarly, Meng et al.~\cite{terrainnet} utilize semantic and geometric data for fast terrain modeling for off-road navigation of a buggy.

\subsection{Semantic Traversability}
Recent approaches to traversability use semantic segmentation to label and assign navigation costs to the observed environment~\cite{pmlr-v164-shaban22a, vstrong, viplanner, stamatopoulou2024dippest}. However, semantic segmentation is often limited to a predefined set of semantic classes and requires extensive data annotation. Despite these challenges, EVORA~\cite{evora} successfully navigates challenging terrain by combining elevation with semantic information and learning the correlation between terrain traction and its uncertainty, resulting in a risk-aware trajectory planner. Ewen et al.~\cite{these_maps} uses semantic segmentation to map material properties, such as friction, as a measure of the cost of traversability, allowing a quadruped robot to walk through snow and ice. RoadRunner~\cite{roadrunner} demonstrates a dune buggy performing high-speed off-road navigation using a traversability segmentation model similar to~\cite{mmseg}. Notably, Kim et al.~\cite{koreaego} use SAM~\cite{sam} and SegFormer~\cite{segformer} to recognize desirable terrain.

\subsection{Self-Supervised Traversability}
Self-supervision has emerged as a pivotal approach in the training of navigation models, enabling robots to learn traversability and interactive behaviors from unlabeled data, thus improving adaptability and robustness in dynamic environments. Kahn et al.~\cite{BADGR} overcome geometric navigation limits by learning physical affordances from real-world experiences, enabling navigation through misleading obstacles such as tall grass and improving autonomously by accumulating data. Cai et al.~\cite{pietra} integrate physics priors into self-supervised models, addressing uncertainties with novel terrains through a balanced, physics-informed approach. They demonstrate improved accuracy and navigational performance in diverse environments, particularly with significant terrain variations. Schoch et al.~\cite{in-sight} propose a self-supervised framework that dynamically interacts with obstacles, integrating traversability scores into semantic maps for complex environments. Validated in simulations and real-world deployments on the robot quadruped ANYmal, it demonstrates effective interactive navigation with promising sim-to-real transfer capabilities.
Following~\cite{in-sight}, which demonstrate that sim-to-real transfer can produce robust models capable of effectively interpreting real-world environments, we apply a similar strategy of using synthetic data to enhance our approach.

\begin{figure*}[ht]
    \vspace{0.25cm}
    \centering
    \includegraphics[width=1.0\linewidth]{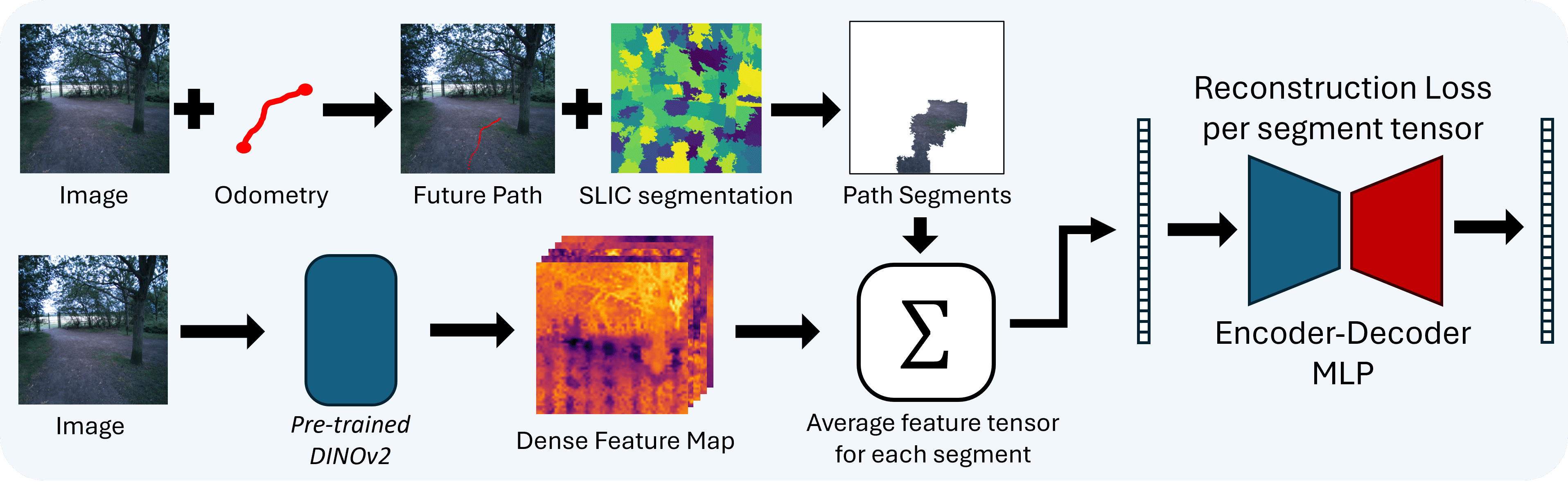}
    \caption{The full pipeline from creating training data to training model as well as the pipeline for \algoname\ at inference.}
    \label{fig:2}
\end{figure*}

\subsection{Anomaly Detection}
Anomaly detection has been popularly adopted for terrain estimation to reduce the cost of mapping unstructured terrain~\cite{scate, roadrunner}. Positive unlabeled (PU) learning, a semi-supervised technique, is particularly effective in scenarios where only a small subset of positive examples is labeled, and a large pool of unlabeled data may contain both normal and anomalous examples. 
Works such as~\cite{scate} use self-supervision and anomaly detection for traversability estimation, using PU data to reduce the need for manual annotation. By incorporating uncertainty quantification, they extract valuable insights from unlabeled or partially labeled data, enabling more efficient learning from large-scale datasets.

\section{METHOD}
Given an RGB image from the robot's egocentric perspective, our goal is to design a model that estimates terrain traversability by generating a costmap for local path planning. Towards the goal, the robot should be capable of avoiding obstacles and non-traversable areas. The re-projected surroundings should clearly differentiate what is traversable or not. The system deployed onto any robot should seemlessly integrate with the robot's local path planner.

We introduce \algoname, a framework for Semantic Traversability Estimation that uses features extracted from future Pose Projections in images. Fig.~\ref{fig:2} illustrates the general information flow in our system. \algoname\ is trained on large amounts of offline data. The data consists of egocentric videos of humans walking across both structured and unstructured terrains, as well as synthetically generated data from simulations of a wide range of environments. \algoname\ learns to recognize terrain features and estimates dense traversability scores from the observed environment. These scores are then projected onto the 3D space around the robot and fed to a local path planner.

\subsection{Data Collection And Pose Projection}
\begin{figure}[ht]
    \vspace{0.25cm}
    \centering
    \includegraphics[width=1.0\linewidth]{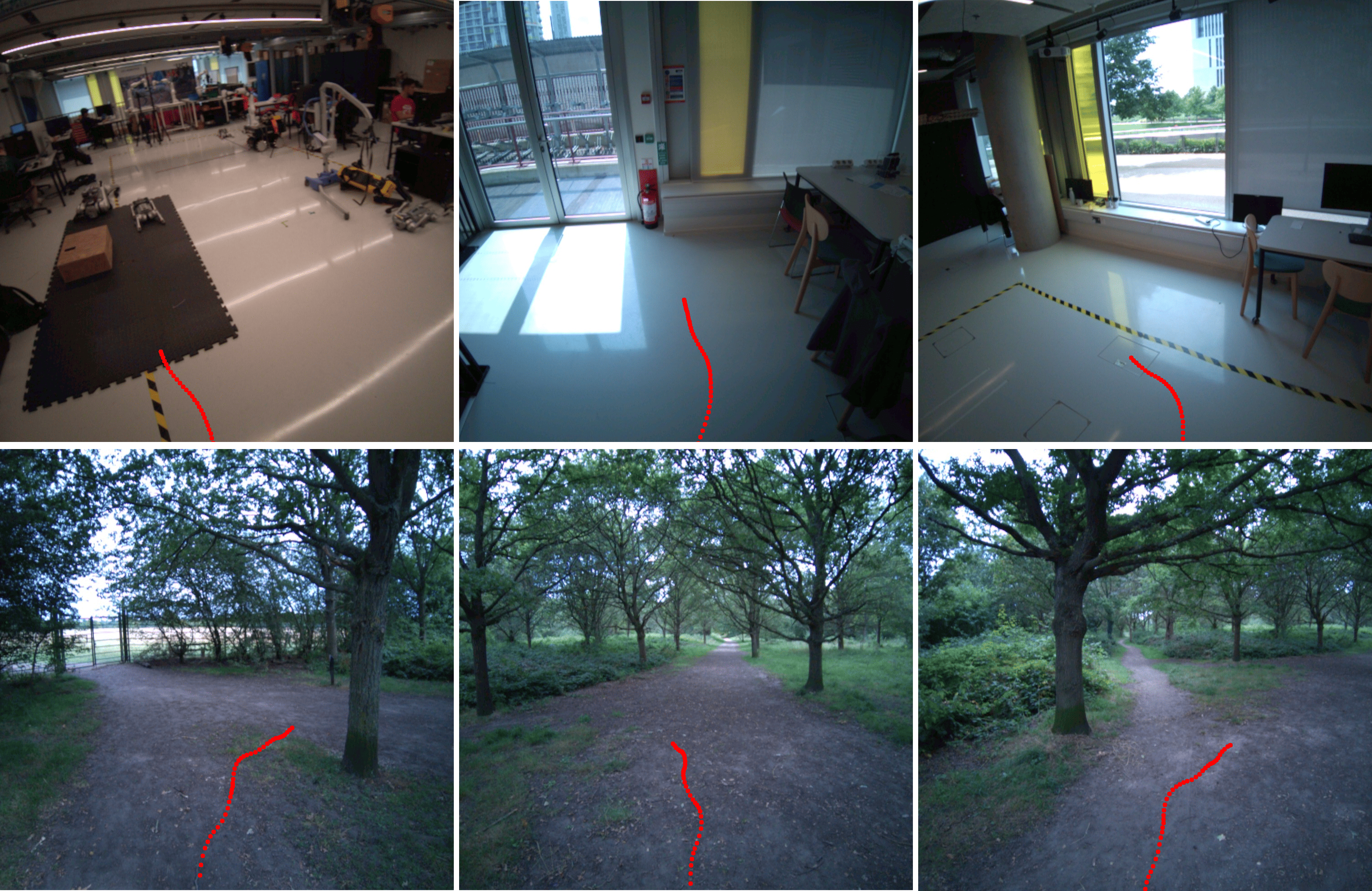}
    \caption{Samples of the data recorded from walking in different environments used with pose projected path on it.}
    \label{fig:3}
\end{figure}
We collect egocentric videos obtained by our custom camera-LiDAR rig, that can be handheld or placed onto a quadruped robot. This allowed capturing the egocentric perspective of a human walking on various terrains. To ensure a more robust terrain estimation, we exclude terrains with large height variations. To label the path trajectories in these data, we use SLAM~\cite{fastlio} to backproject the trails onto the captured RGB images. Human-recorded data with future projected path trajectories can be seen in Fig.~\ref{fig:3}. This is close to the way~\cite{Where_Should_I_Walk, wvn} used to avoid manual labeling. To obtain as much training data as possible, we project future paths onto every recorded image.

To project future path trajectories onto an image, we first define the trajectory of the rig's odometry in world frame as 
\[
\pazocal{P} = \{ \mathbf{T}_i \}_{i=0}^{n}
\]
where each \( \mathbf{T}_i \in SE(3) \) is the pose of the device at time step \( i \), consisting of a rotation matrix \( \mathbf{R}_i \in SO(3) \) and a translation vector \( \mathbf{t}_i \in \mathbb{R}^3 \):
\[
\mathbf{T}_i = \begin{bmatrix}
\mathbf{R}_i & \mathbf{t}_i \\
\mathbf{0}^\top & 1
\end{bmatrix}
\]
The odometry trajectories are recorded in the world frame, but to project future trajectories into each image, we must first transform the points from the world frame to the device (camera) frame. Given a point \( \mathbf{X} \in \mathbb{R}^3 \) in the world frame, its position in the device frame at time \( i \) is:
\begin{equation}
\mathbf{x}_i = \mathbf{T}_i^{-1} \mathbf{X}
\end{equation}
To project the future poses onto the ground, we account for the device’s height \( H \) by translating the points along the z-axis in hte world frame for gravity alignment:
\begin{equation}
\mathbf{x}_i^{\text{ground}} = \begin{bmatrix} x_i \\ y_i \\ z_i - H \end{bmatrix}
\end{equation}
These ground plane points are then transformed back into the device frame using the inverse pose \( \mathbf{T}_i^{-1} \).

Finally, to obtain the pixel coordinates in the image, we project the 3D points \( \mathbf{x}_i \) from the device frame onto the 2D image plane using the camera’s intrinsic matrix \( \mathbf{K} \):
\begin{equation}
\mathbf{p}_i = \frac{1}{z_i} \mathbf{K} \begin{bmatrix} x_i \\ y_i \\ z_i \end{bmatrix}
\end{equation}
where \( \mathbf{p}_i = [u_i, v_i]^\top \) represents the pixel coordinates. To avoid including terrain that has not been walked on, we project only the next $40$ future trajectory poses to avoid overlap during turns or poses close to the horizon of an image.

\subsection{Synthetic Data Generation}
\begin{figure}[ht]
    \vspace{0.25cm}
    \centering
    \includegraphics[width=1.0\linewidth]{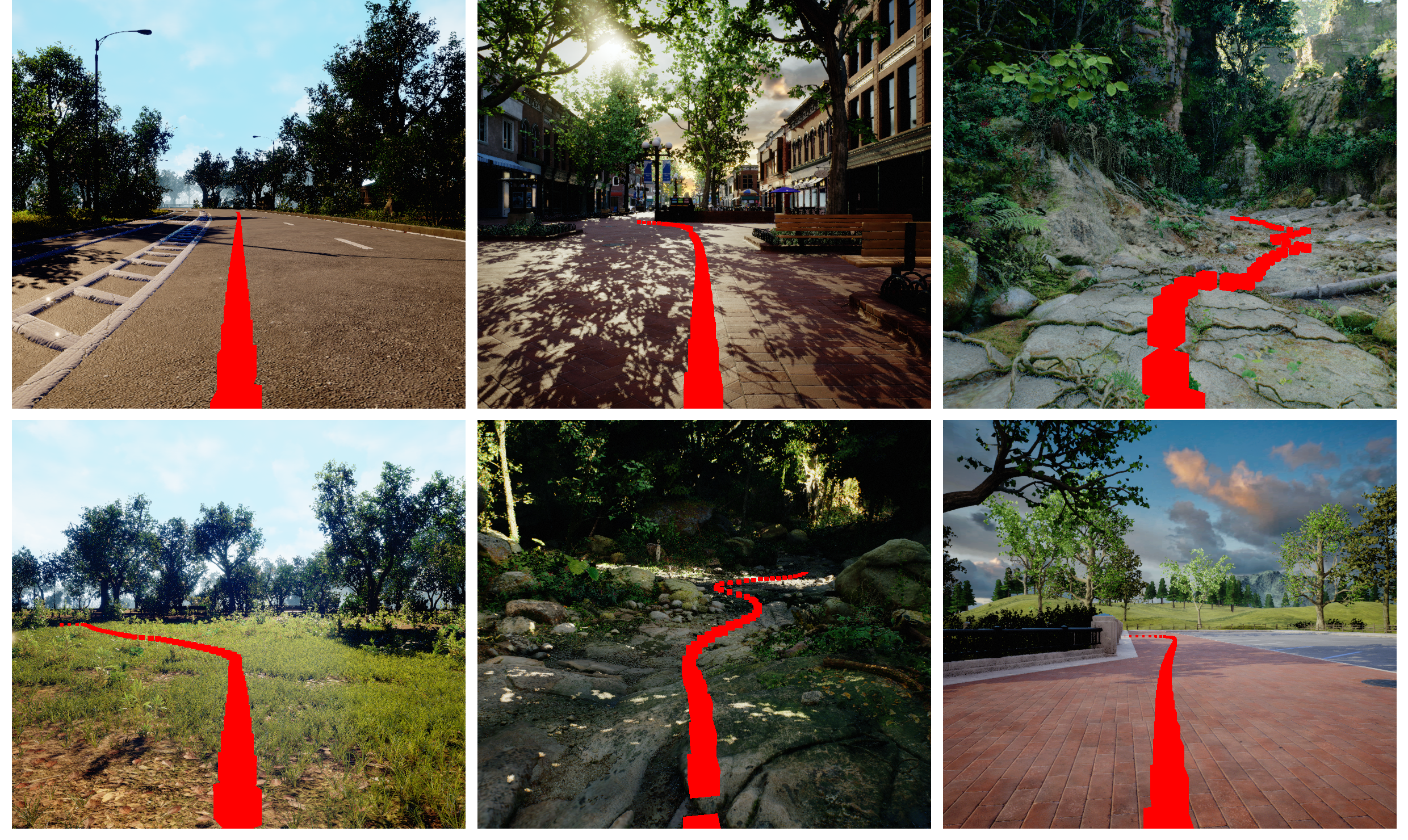}
    \caption{Unreal Engine simulation environment data used for training. Red line shows the trajectory the simulation camera takes projected onto the ground.}
    \label{fig:4}
\end{figure}
We developed a plugin for Unreal Engine~\cite{unrealengine} (UE) that simplifies the generation of paired image and future-path trajectory data using a simulated rig within any UE environment. The plugin generates RGB images along with corresponding ground-truth trajectory data mapped to the pixels in each image, as well as full IMU trajectory data. A custom C++ Actor can be instantiated in any level, alongside a spline that can be adjusted to create human-like trajectories. The spline is automatically adjusted by projecting it onto the terrain, with its height set at a fixed distance above the ground. Data are then collected along this spline at adjustable velocities. Fig.~\ref{fig:3} illustrates six examples of the various simulation environments used for feature extraction. This plugin allows us to generate large and varied datasets that would otherwise be prohibitively time-consuming and resource-intensive to collect in real-world settings.

\subsection{Segmentation And Feature Extraction}
After processing the recorded odometry and image data, the list of pixels for each image is then used to extract the ground on which it was walked. For fast segmentation of what ground the pixel in each image land on, we use a variation of the SLIC~\cite{slic} geometric-based superpixel clustering segmentation algorithm. We chose to segment the images with the SLIC set to $400$ superpixels and the compactness of $15$ to accommodate enough detail in each segmentation. The resultant SLIC segmentation is a mask of segments, each with their own unique value.

To extract the features to learn from the given image, we used the backbone of the self-supervised DINOv2 Vision transformer~\cite{dinov2, hadjivelichkov2022one} from Meta. For this project, we chose to use the smallest backbone of the DINOv2 model to achieve the fastest inference time, which outputs an image with 384-dimensional pixel-wise feature embeddings. As the output of the DINOv2 backbone is the input of the image divided by the patch size used of $14$, the input of the image was set to $700\times700$, giving an output of $50\times50$. The value of $700$ was chosen to reduce processing costs without losing too much information from each image. This results in a $50\times50\times384$ dense image feature output.

The SLIC segmented image is then downsized to the same size as the DINOv2 output of $50\times50$ using nearest-neighbor interpolation to retain the same pixel values after resizing. From these two originally sized and downsized segmentation masks, we map the values of the segments that contain the projected pose pixels on the original image to the same segment value on the resized image. From this mapping, we can accurately choose the pixels in the dense DINOv2 output that correlate to the future pose projection pixels on the original image.The remaining pixels not on the projected path are set to zero.

Given the downsized SLIC segmented mask that only highlights the specific pose projection region of interest within the image, we extract the feature embeddings corresponding to the pixels within the masked area. For each of the $384$ dimensions, we calculate the mean value of the feature in the selected pixels, resulting in a single 384-dimensional feature vector, $\mathbf{f}$, which represents the average embedding for the masked region.

The superpixel feature averages are computed using PyTorch's $scatter\_reduce$ operation, which enables efficient summation of pixel-wise embeddings across the segmented regions. This operation significantly reduces the computational complexity of averaging over the large number of pixels within the mask, providing a scalable solution. 
We also utilize mixed precision for DINOv2, reducing its input image from a 32-bit floating point (FP32) to a 16-bit floating point (FP16) to speed up the inference of the neural network.

\subsection{Inference}
To recognize the traversed segments, we aim to learn the distribution of all the averaged features, $\mathbf{f}_{n}$, from the traversed segments. For this purpose, we deploy a Multi-Layer Perceptron (MLP) neural network model to learn to recognize each tensor. The MLP is designed as a fully connected classical feedforward network with hidden $7$ layers, where each layer contains $[256,128,64,32,64,128,256]$ units. The number of hidden layers was chosen empirically as more hidden layers did not show signs of better feature recognition, and fewer hidden layers did not capture enough information. We use ReLU activation function between layers to ensure non-linearity. The MLP is structured as an encoder-decoder architecture, where the encoder compresses the input $\mathbf{f}_{n}$ into a lower-dimensional latent space, and the decoder reconstructs $\mathbf{f}_{n}$ to its original size.

The reconstruction loss is set to be the mean squared error (MSE) between the reconstructed $\mathbf{f}_{n}$ and the original input $\mathbf{f}_{n}$ across all dense dimensions:
\begin{equation}
    \pazocal{L}_{\text{rec}} = \frac{1}{N} \sum_{n=1}^{N} \| \mathbf{f}_n - \hat{\mathbf{f}}_n \|^2,
\end{equation}
where $N$ denotes the number of dimensions of the dense feature embedding, $\mathbf{f}_{n}$ denotes the input tensor, and $\hat{\mathbf{f}}_{n}$ denotes the reconstructed tensor. This design allows the MLP autoencoder to learn the feature representations of the terrain segments seen during training. Consequently, the network reconstructs previously traversed terrain with a low reconstruction loss, whereas unseen terrain (e.g., a tree or a large boulder) results in a higher loss, facilitating anomaly detection.

\subsection{Integration}
We used the CMU navigation stack~\cite{tare} as our local path planner. To integrate the output of \algoname\ with the local planner, we projected the traversability cost into the 3D space using a depth camera. \algoname\ inference pointcloud to fit the height-based value range of the Falco~\cite{Falco} local planner. To enable the robot system to maneuver dynamically through tall vegetation and other obstacles close to the camera, a minimum distance of $2$m from the camera to the object was used, ensuring smooth path planning despite larger momentary camera occlusions. 

\section{VALIDATION AND EXPERIMENTS}\label{Sec:exp}
\subsection{Model Accuracy}
\begin{figure}[t!]
    \vspace{0.25cm}
    \centering
    \includegraphics[width=1.0\linewidth]{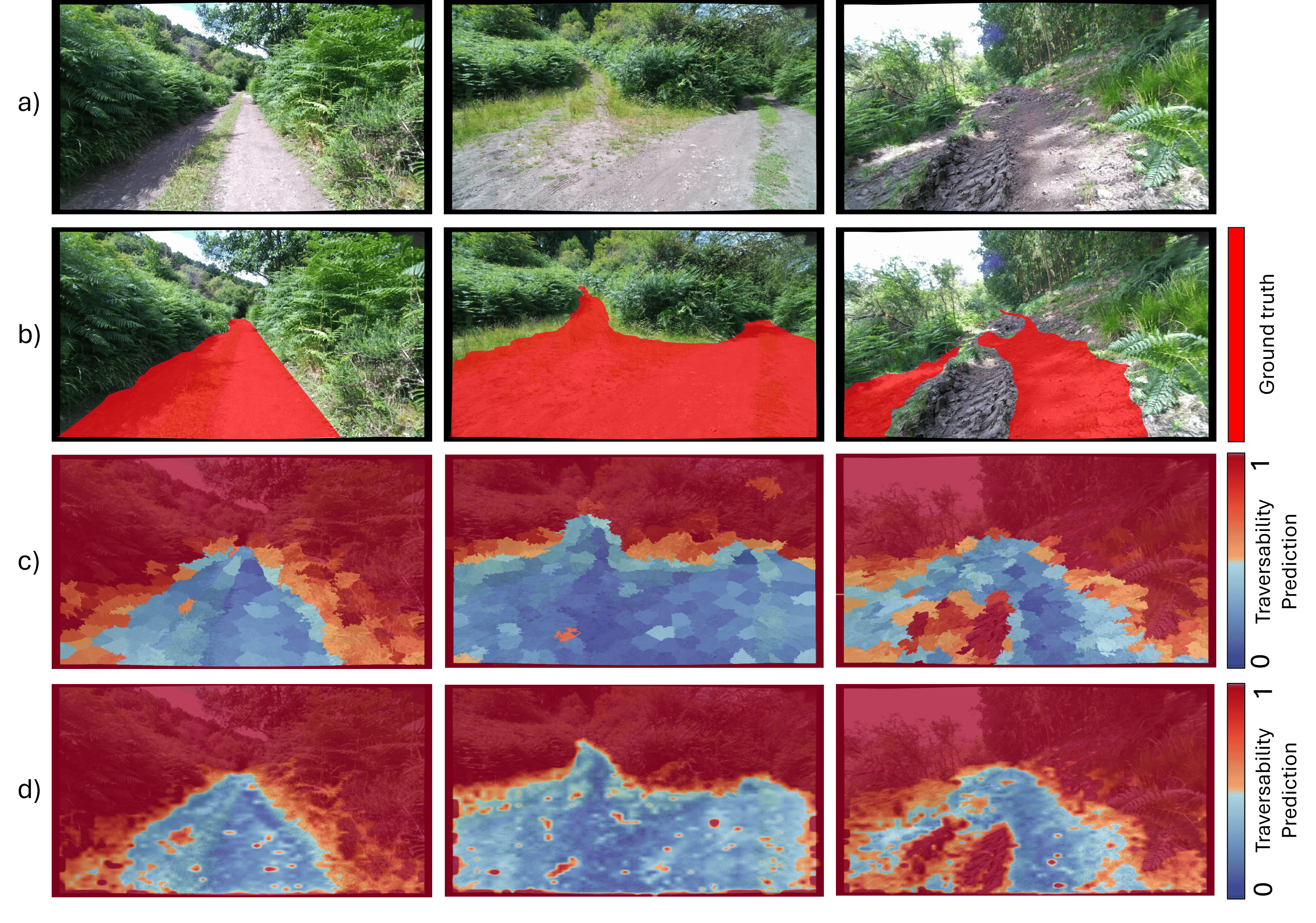}
    \caption{\algoname\ reconstruction loss on unseen forest data. a) sample environment, b) traversability ground truth, c) \algoname\ traversability cost feature segment wise, d) \algoname\ traversability cost feature pixel wise. Segments and pixels that are the darkest red is considered not traversable.}
    \label{fig:5}
\end{figure}

To validate the accuracy of \algoname\ estimated terrain traversability, we manually annotated $500$ unseen images of a walk in a forest. We used this as ground truth to compare with the output of our introduced system. Fig.~\ref{fig:4} shows the ground truth traversability and the output \algoname, both segment-wise and pixel-wise feature. The loss of output reconstruction is capped at a maximum value of $10$ and then normalized within the range $[0,1]$. This made it possible to set a threshold for a lower-bound reconstruction loss that would signify fully traversable terrain. We used the $500$ labeled ground-truth traversability regions to fine-tune the traversability threshold. The optimal threshold with the traversability coverage closest to the ground truth was set to $0.35$. Segments or pixels below this threshold are considered traversable. 

In Table~\ref{tab:1} we show the accuracy of \algoname\ trained int data configurations. The threshold for all accuracy tests was empirically adjusted to give the highest accuracy for each dataset.
As intuitively expected, models trained on real-world forest data perform best on a forest-based validation dataset. Indoor laboratory data was shown to be unable to generalize to the outdoor environment and hinder model accuracy when combined with forest training data. Finally, we note that the data from simulated environments is somewhat transferable to real environments (with an accuracy of $0.684$), and it enhances overall performance when combined with forest data. The comparison shows that the model performs best when all relevant training data is used. The size is the number of images in the dataset.

\captionsetup{font=small}
\begin{table}[ht]
    \vspace{2mm}
    \centering
    \caption{Model accuracy when trained on various data sources.}
    \begin{tabular}{p{1.5cm}p{2.5cm}p{1.5cm}>{\centering\arraybackslash}m{1.75cm}}
    \toprule
        Data type &
        Data configuration & Dataset size & Accuracy[\%]\\
        \hline
        Real & Forest & $55,580$ & $0.803$\\
        Real & Indoor lab & \hspace{3.4pt}$5,384$ & $0.438$\\
        Real & Forest + Indoor lab & $60,964$ & $0.769$ \\
        Simulated & UE Environments & $26,954$ & $0.684$\\
         \hline
        Mixed & All data combined & $87,918$ & \textbf{0.835}\\
    \bottomrule
    \end{tabular}\\
    \vspace{0.1cm}
    UE = Unreal Engine.
\label{tab:1}
\end{table}

\subsection{Real-World Traversability-based Legged Path Planning}
In the real world, the ANYbotics ANYmal-D legged robot was used. We equipped the robot with our own custom sensor rig containing a Zed2 stereo camera, a Livox Mid360 LiDAR, an Intel Nuc 11, and a Nvidia Jetson Orin AGX. During all experiments, the ZED2 stereo camera provided the RGB input of \algoname\ along with its depth image to project the traversability scores in 3D. The Livox LiDAR along with the Intel NUC ran SLAM for odometry, while the NUC also ran the CMU navigation stack's~\cite{tare} local Falco planner~\cite{Falco}.  Orin ran the python-implied inference node of \algoname\ using pytorch~\cite{pytorch} and ROS. More experimental videos and detailed model demonstrations are available on the project’s website:\\ \url{https://rpl-cs-ucl.github.io/STEPP/} 

\subsubsection{Indoor Environments}
\begin{figure}[t]
    \vspace{0.25cm}
    \centering
    \includegraphics[width=1.0\linewidth]{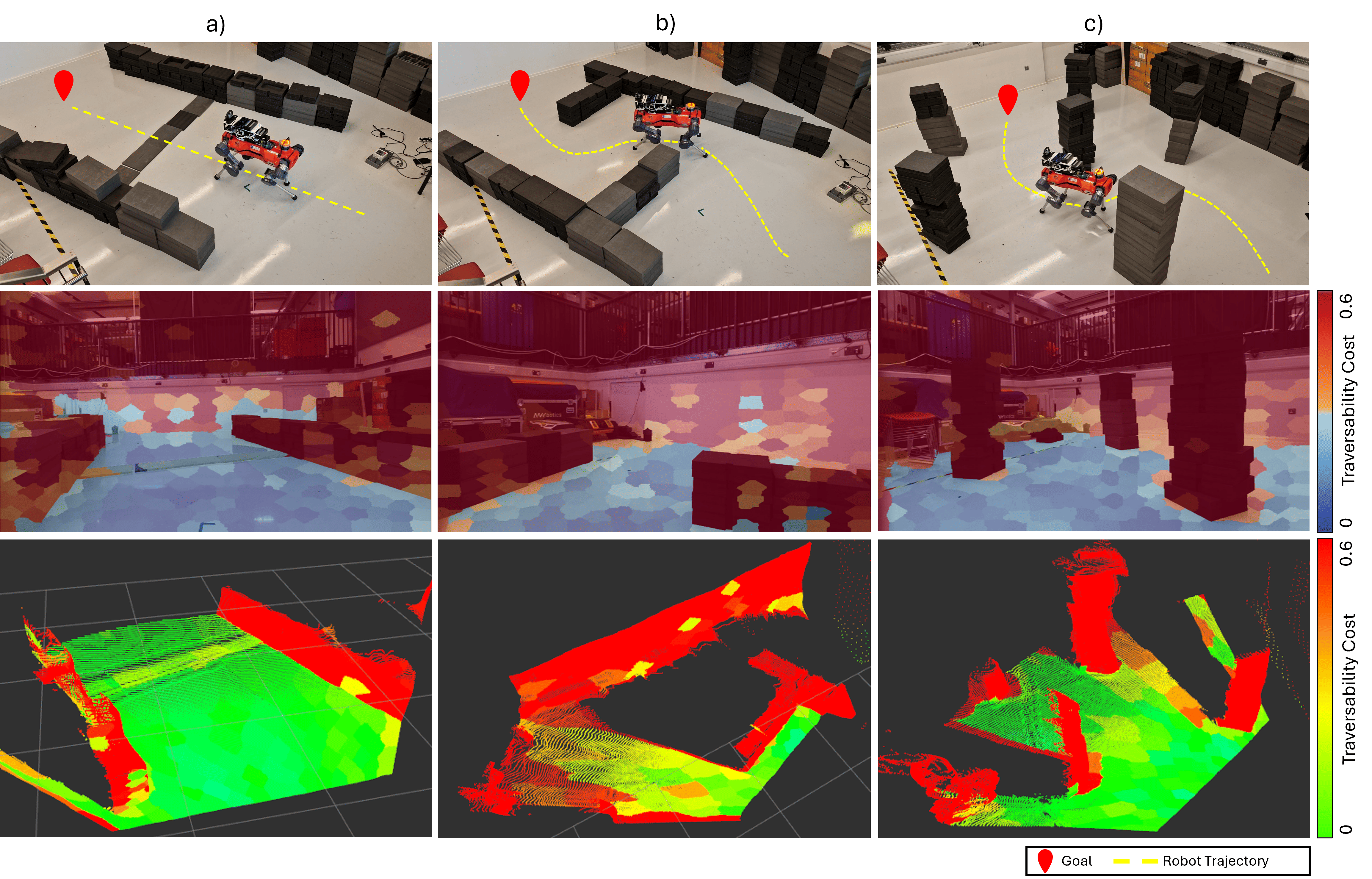}
    \caption{Pictures of three indoor maze environments with overlaid robot trajectories toward a goal (top) with traversability cost predictions in the robot's egocentric perspective (middle), and projected pointcloud for navigation (bottom). Lower traversability cost indicates the terrain is more traversable.}
    \label{fig:6}
\end{figure}

To assess the performance of our traversability estimator \algoname, we conducted a series of indoor experiments that featured navigation through maze environments and obstacle courses. These experiments were set up to mimic an indoor environment with \algoname\ trained in similar rooms. The mazes included a variety of obstacles and configurations to test the robot's ability to interpret and adapt to new and unforeseen environmental contexts, thereby evaluating the robustness of our proposed methods when faced with structured terrain. Fig.~\ref{fig:6} shows three indoor experiment scenarios along with the robot camera view overlayed with the traversability cost at the shown position and the projected point cloud. In all mazes, the robot was able to navigate through, as long as the local planner allowed it with the size constraints. In particular, the model learned to walk across flat black mats laid on the floor without being specifically told to do so. In addition, the height-based terrain estimator~\cite{tare} managed to walk through all maze configurations.

\subsubsection{Outdoor Environments}
\begin{figure}[t]
    \vspace{0.25cm}
    \centering
    \includegraphics[width=1.0\linewidth]{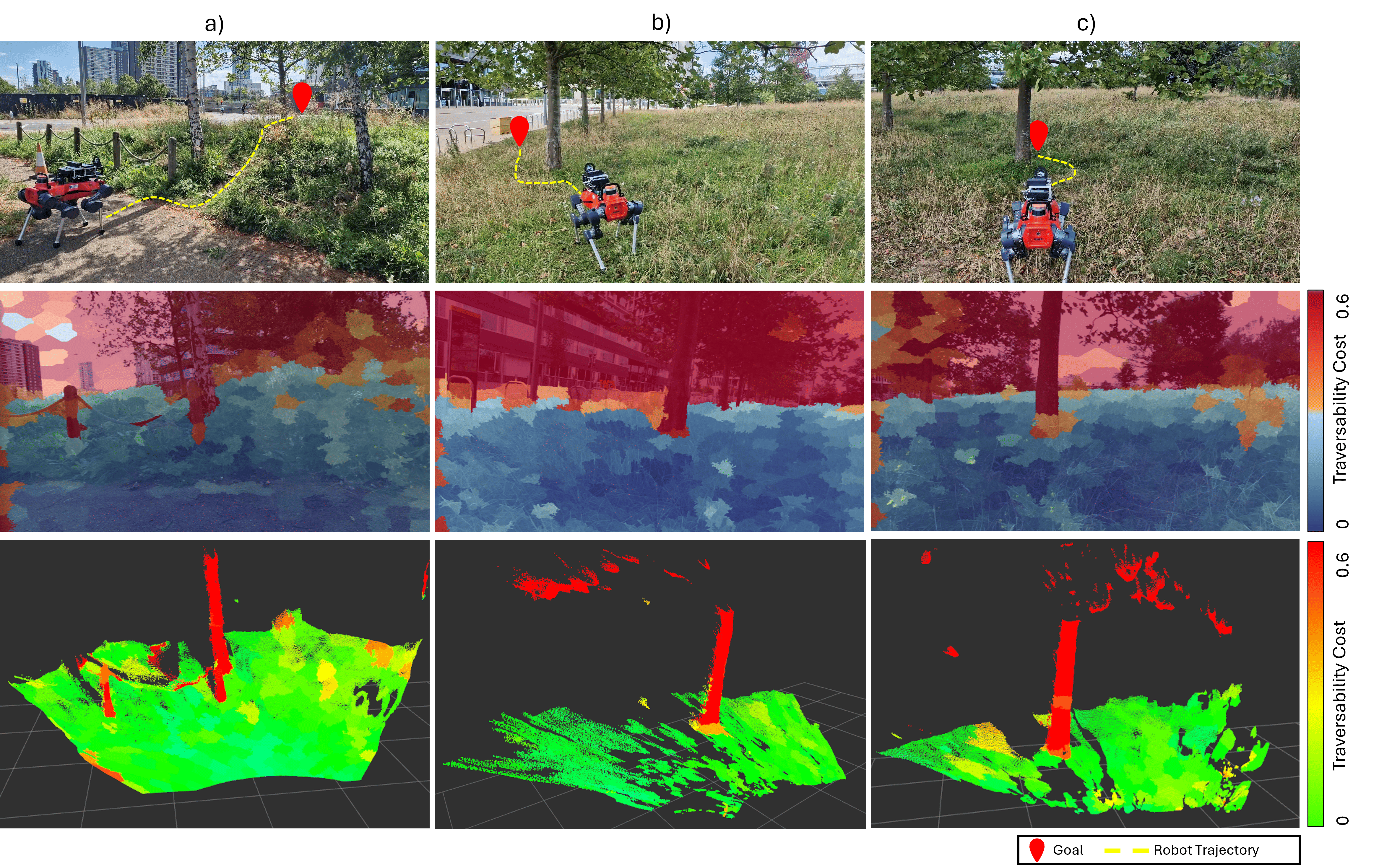}
    \caption{Pictures of three outdoor environments with overlaid robot trajectories toward a goal (top) with traversability cost predictions in the robot's egocentric perspective (middle),  and projected pointcloud for navigation (bottom). Lower traversability cost indicates the terrain is more traversable.}
    \label{fig:7}
\end{figure}

We conducted further evaluations of \algoname\ on unstructured terrain through a series of outdoor experiments in forest-like environments. These experiments were designed to assess the model's capacity to distinguish between traversable and non-traversable vegetation. Specifically, it was tested on its ability to navigate through short and tall grass while avoiding trees and larger obstacles. Unlike conventional height-based terrain analysis methods, such as~\cite{tare}, \algoname\ successfully navigated through tall grass and successfully avoided taller vegetation and substantial obstacles, demonstrating its enhanced perceptual and navigational capabilities in complex outdoor settings. Fig.~\ref{fig:7} shows three experiment scenarios in outdoor environments --- in the first experiment (a), the robot avoids the rope fence and the tree and walks up the ridge through tall grass, shown in the projected cost map. The next two experiments, (b) and (c), show the robot walking through short and medium-tall grass and avoiding trees. In the cost map, the ground and short plants are shown to be traversable (in green), while obstacles such as the fence and trees are correctly identified (in red) on the cost map. We attempted navigating in these same scenarios using ~\cite{tare} -- the method failed in scenario (a) and did not walk smoothly across the ground in scenarios (b) and (c). 

\section{DISCUSSION AND CONCLUSION} \label{Sec:concl}
We propose a model and framework to perform traversability estimation leveraging pose-projected features and deploy it in real-world robot systems. Our method successfully integrates dense pixel-wise embedding with spatial transformations to predict and evaluate terrain navigability in real-time. The experimental results, both indoor and outdoor, demonstrate performance in diverse environments and validated the key idea behind our approach of utilizing the latent space information priors from pre-trained foundation models, enabling generalization and adaptation in unseen scenarios when only given positively labeled data.

We have demonstrated that the \algoname\ model can be deployed to navigate various terrains out of the box without further learning required. However, there are few limitations that warrant further discussion:
\begin{itemize}
    \item The robot infers at $2.5$ Hz, leading to delayed responses to scene changes due to synchronization lag between odometry and depth projection. This can compromise the effectiveness in real time. A faster inference of features would improve this.
    \item \algoname\ shows variability in traversability cost estimates for similar images, reflecting the uncertainty of the output that can affect the reliability of navigation. 
    \item Although \algoname\ correctly estimates the traversability of terrain it has seen during training, it struggles with terrain types or scenarios absent from the training data, often misidentifying or misrepresenting unfamiliar environments. This could be countered by a larger dataset or a different reconstruction network. 
    \item The use of SLIC for fast image segmentation can inaccurately emphasize shape and cluster size, sometimes erroneously dividing or grouping objects within segments. Currently, there is no more detailed segmentation model with the same speed as SLIC does not exist, but it would improve the accuracy of \algoname. 
    \item Since the data gathered are ego-centric, the walkable terrain tends to be in the lower middle of the images. This leads to model bias towards that region of the images and may cause incorrect predictions in the rest of the image. Data augmentations could be sufficient to mitigate this limitation.
    \item The effectiveness of \algoname\ in the real world deployment is highly dependent on the precision of the depth sensing. Calibrating the cameras to a more accurate LiDAR would increase the accuracy of the model.
\end{itemize}


\vspace{4mm}
\bibliographystyle{IEEEtran}
\bibliography{icra2025seb}

\end{document}